\documentclass[dvipsnames,format=sigconf,anonymous=false,review=false,nonacm]{acmart}

\usepackage{subfig}
\usepackage[noend]{algorithmic}
\usepackage{algorithm}
\usepackage{fancyvrb}

\AtBeginDocument{%
  \providecommand\BibTeX{{%
    \normalfont B\kern-0.5em{\scshape i\kern-0.25em b}\kern-0.8em\TeX}}}

\setcopyright{none}

\begin{document}

\title{Inexact Simplification of Symbolic Regression Expressions with Locality-sensitive Hashing}

\author{Guilherme Seidyo Imai Aldeia}
\orcid{0002-0102-4958}
\affiliation{%
  \institution{Federal University of ABC}
  \city{Santo Andre}
  \state{São Paulo}
  \country{Brazil}
}
\email{guilherme.aldeia@ufabc.edu.br}

\author{Fabrício Olivetti de França}
\orcid{0000-0002-2741-8736}
\affiliation{%
  \institution{Federal University of ABC}
  \city{Santo Andre}
  \state{São Paulo}
  \country{Brazil}
}
\email{folivetti@ufabc.edu.br}

\author{William G. La Cava}
\orcid{0000-0002-1332-2960}
\affiliation{%
  \institution{Boston Children's Hospital}
  \institution{Harvard Medical School}
  \city{Boston}
  \state{Massachusetts}
  \country{USA}
}
\email{william.lacava@childrens.harvard.edu}

\begin{abstract} 
Symbolic regression (SR) searches for parametric models that accurately fit a dataset, prioritizing simplicity and interpretability.
Despite this secondary objective, studies point out that the models are often overly complex due to redundant operations, introns, and bloat that arise during the iterative process, and can hinder the search with repeated exploration of bloated segments.
Applying a fast heuristic algebraic simplification may not fully simplify the expression and exact methods can be infeasible depending on size or complexity of the expressions.
We propose a novel agnostic simplification and bloat control for SR employing an efficient memoization with locality-sensitive hashing (LHS).
The idea is that expressions and their sub-expressions traversed during the iterative simplification process are stored in a dictionary using LHS, enabling efficient retrieval of similar structures. 
We iterate through the expression, replacing subtrees with others of same hash if they result in a smaller expression. 
Empirical results shows that applying this simplification during evolution performs equal or better than without simplification in minimization of error, significantly reducing the number of nonlinear functions.
This technique can learn simplification rules that work in general or for a specific problem, and improves convergence while reducing model complexity.
\end{abstract}




\maketitle

\section{Introduction}
Symbolic regression (SR) addresses the challenge of jointly optimizing the parameters and structure of a function.
Given a set of $d$-dimensional inputs $\mathcal{X}$ and target outputs $\mathbf{y}$, it solves $\mathbf{y} \approx \mathbf{\hat{y}} = \hat{f}(\mathcal{X}, \mathbf{\hat{\theta}})$, typically by minimizing the difference between $\mathbf{y}$ and $\mathbf{\hat{y}}$ while concurrently prioritizing simplicity \cite{contemporarySRMethods}.
Since first proposed by Koza as an application of genetic programming (GP) \cite{koza1994genetic}, it has been successfully used in several fields, such as clinical decision support \cite{la_cava_flexible_2023}, financial modeling \cite{math11092108}, aerospace engineering \cite{LACAVA2016892}, and physical sciences \cite{Angelis2023}.

Despite its success, SR is an NP-hard problem \cite{SymbolicRegressionIsNPHard}, meaning that searching for the best regression model is computationally inefficient unless P=NP is proved.
The search space is vast, containing isomorphic mathematical expressions \cite{burlacu_hash-based_2021}, introns (parts of the model that do not influence predictions) \cite{affenzeller2014gaining}, bloat (growth in model size with unjustified improvement in loss) \cite{luke_comparison_2006}, and model overparameterization (excessive number of parameters to tune) \cite{10.1145/3583131.3590346}.
While these do not directly affect model accuracy, they can inflate model size, decreasing simplicity and interpretability.

There are several different approaches in the literature for tackling this problem, such as automatic simplification of models \cite{helmuth_improving_2017}, parametric rules to rewrite expressions \cite{kulunchakov_creation_2017}, Bayesian loss metric for model selection \cite{bomarito_bayesian_2022}, constrained search space by restricting model structure \cite{10.1162/evco_a_00285}, or application of an exact simplification procedure such as equality saturation~\cite{10.1145/3583131.3590346}.
Although all of those have their own benefits, they also have limitations, chiefly the additional computational cost and the need to manually write the algebraic simplification rules.

We propose a technique to dynamically build the simplification rules to address these limitations by memoizing observed equivalences with hashing.
In short, we build a hash table where the key is created using locality-sensitive hashing (LHS) \cite{gionis_similarity_nodate} based on the prediction vector, and the value is the smallest observed subtree corresponding to that vector.
LHS is often used to efficiently retrieve nearest-neighbor points in a database system \cite{jafari_survey_2021}.
By analogy, our proposal transforms expressions into similar expressions within the phenotype (i.e., behavior) space that are smaller.

When integrating and applying the simplification into every expression throughout the generations, we observed benefits such as faster convergence and reduced mean squared error.
The size of the returned expression remained the same, but its complexity was significantly reduced when considering the use and chaining of nonlinear functions.
Furthermore, we analyzed the expressions identified as equivalent and noticed that the algorithm successfully captures many known algebraic identities.

The paper is organized as follows.
Section \ref{sec:relatedwork} provides an overview of related work in bloat control, hashing, and simplification methods in SR.
Section \ref{sec:lsh} introduces the application of Locality-Sensitive Hashing (LSH) in our simplification method.
Section \ref{sec:memoization} details simplifying expressions through memoization with LSH.
Section \ref{sec:methods} outlines the experimental methods, including the SR framework used to test the method.
Section \ref{sec:results} presents and discusses empirical findings, focusing on the impact of our approach on solution quality, expression size, and eliminating unnecessary operations.
Finally, Section \ref{sec:conclusions} draws conclusive remarks, summarizing key contributions and suggesting future work.

\vspace{4.0em}

\section{Related work}~\label{sec:relatedwork}

Helmuth et al. \cite{helmuth_improving_2017} proposed an automatic simplification in the program synthesis context by randomly removing subtrees and accepting the change if that did not affect the final prediction.
Burlacu et al. \cite{burlacu_hash-based_2021} explored the idea of hashing an SR tree using some basic algebraic rules for handling commutative operations, such that two equivalent subtrees are hashed to the same value.
Nguyen and Chu explored indirect simplification through mutation \cite{nguyen_semantic_2020}, pruning a population percentage by replacing subtrees with a newly small random tree with similar semantics.

Some previous work focuses on sets of rules to perform simplification.
Kulunchakov \cite{kulunchakov_creation_2017} defined equivalent algebraic models as those that produce the same outputs over a small range and by evolving expressions with GP to use as replacements.
In  \cite{10.1145/3583131.3590346}, de França and Kronberger proposed using equality saturation, an enumerative simplification algorithm, to remove redundant parameters of symbolic expressions. They noticed that overparametrization and unnecessary complexity are predominant in current SR algorithms.
Some attempts rely on something other than the manipulation of symbols.
Bomarito et al.~\cite{bomarito_bayesian_2022} proposed to reduce bloat by using a custom Bayesian fitness metric with regularization.

Contrary to previous work, our approach does not rely on sets of rules or expensive computing resources. We propose a data-driven and computationally efficient way of inexact simplification that can be embedded into any GP framework, being agnostic because it only requires slicing the expressions and replacing subtrees.

\section{Locality-Sensitive Hashing}~\label{sec:lsh}

\begin{figure}[tbh]
    \centering
    \includegraphics[width=.85\linewidth]{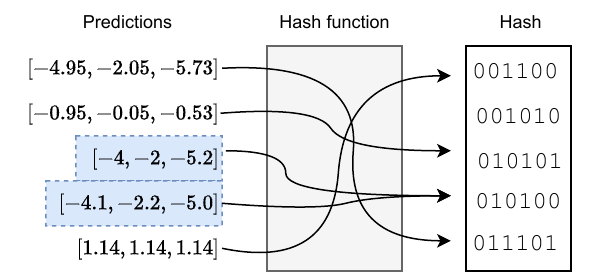}
    \caption{Given a set of vectors (\textit{i.e.} the predictions), similar predictions are mapped to the same hash. The two highlighted vectors present close values in this example and are mapped into the same hash.}
    \label{fig:lsh_mapping}
\end{figure}

Locality-sensitive Hashing (LSH) was first introduced in \cite{gionis_similarity_nodate} as a technique designed to efficiently find approximate neighbors in high-dimensional spaces through the usage of hash functions that preserve local proximity \cite{jafari_survey_2021}.
The main idea is to hash input vectors so that similar vectors are more likely to be mapped to the same hash bucket, while very distinct vectors are less likely to do so. Figure \ref{fig:lsh_mapping} illustrates this process.

Here, we employ the SimHash LSH method as a proof of concept~\cite{charikar2002similarity}, as it has a straightforward implementation and requires few lines of code to be incorporated into existing frameworks.
This results in a trade-off between computational efficiency and accuracy, beneficial in scenarios where exact similarity search becomes impractical \cite{jafari_survey_2021} due to the curse of dimensionality \cite{10.1145/502807.502808}.

Given a hash size of $b$ bits and a set of $d$-dimensional data $\{x_i \in \mathbb{R}^d\}$, we first stipulate the hash size as $b$ bits. 
Then, we create a plane $\mathbf{P} \in \mathbb{R}^{b \times d}$ where each 
$\mathbf{P}_{(i,j)} \sim \mathcal{N}(0, 1)$.

Whenever we want to query a data point $x_i$, we first calculate the matrix multiplication $\mathbf{P} \cdot x_i$, resulting in the vector $\mathbf{q} \in \mathbb{R}^{b \times 1}$, and the hash $h(x_i)$ is calculated as

\begin{equation*}
    h(x_i)_j = \begin{cases} 1 & q_j > 0 \\ 0 & \text{otherwise} \end{cases}
\end{equation*}

This forms a bit string that is used as the key to the hash table. The main property of this hash function is that the probability of two hashes being equal, i.e., $Pr[h(x) == h(y)] = 1 - \frac{\theta(x,y)}{\pi}$, is proportional to their cosine similarity, and cosine similarity is proportional to the normalized $l_2$-euclidean distance.

Every hash table entry will store all queried objects with that same hash value.
After inserting many objects, similar objects will be clustered around the same keys. The higher the number of bits used to generate the hash key, the larger the number of less dense clusters, increasing the accuracy of the similarity estimation.

\section{Simplifying expressions by memoization}~\label{sec:memoization}

The idea of inexact simplification consists of having a simplification table where the key is the SimHash (as described in \S\ref{sec:lsh}), and the values are a list of expressions hashed to that particular key.

\begin{figure*}[tbh]
    \centering
    \includegraphics[width=\linewidth]{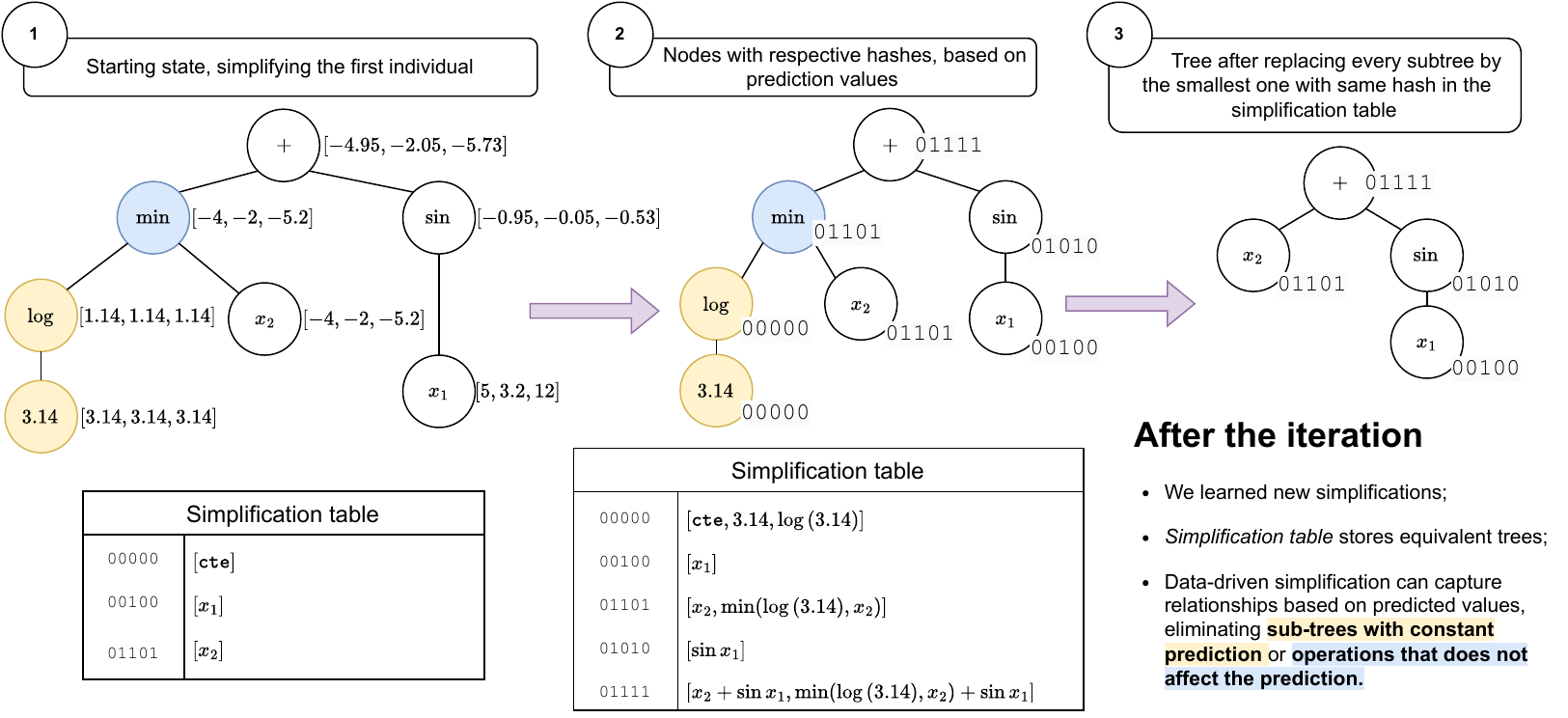}
    \caption{In the first stage ($1$), we have a simplification table with only the problem variables plus the constant. Every node is seen as the root of a subtree and can generate a prediction vector. The second stage ($2$) uses the predictions to get hash values for each node, updating the simplification table. Finally, we get the simplified tree by replacing the nodes with the smallest subtree of the same hash in the simplification table.}
    \label{fig:lsh_simplification}
\end{figure*}

In the first step, we evaluate the expression, keeping a trace of the evaluation at every node. At this point, every node will contain a vector of predictions corresponding to the evaluation of that subtree. In the next step, we traverse the tree again~\footnote{In practice, we only need to traverse the tree once.} hashing these vectors into the binary string key. Finally, if this key contains any element in the hash table and the closest value to this subtree is within a threshold, we replace this subtree with the smallest tree in this table entry. If there is no entry for this key, we create a new entry with this subtree. Figure \ref{fig:lsh_simplification} depicts the proposed algorithm to use LSH to simplify symbolic expressions.

We force every constant vector to have the same hash by setting the prediction to zero if the variance of the vector is zero. This avoids growing the hash table unnecessarily. The threshold also alleviates the fact that nonlinear equations can have non-unique solutions, so they are considered equivalent as long as the predictions are within the threshold.
Algorithm \ref{alg:init_table} describes the initialization of the table, and Algorithm~\ref{alg:hash_simplify} describes the simplification process.

\begin{algorithm}[tbh]
    \caption{initialize\_table}\label{alg:init_table}
    \begin{algorithmic}[1]
        \REQUIRE{ Training data points $(\mathcal{X}^t, \mathbf{y}^t)$, single node constant tree $\texttt{cte}$ }
        \ENSURE{ simplification table $\texttt{st}$ }

        \STATE $\texttt{st} \leftarrow \text{ empty mapping of (key : value)}$

        \FOR{$\texttt{terminal} \in \{\texttt{cte}\} \cup \{x \in \mathcal{X}^t\}$}
            \STATE $\texttt{pred} \leftarrow \texttt{terminal}.\text{predict}(\mathcal{X}^t)$
            \STATE $\texttt{lhs}.\text{index}(\texttt{pred})$
            \STATE $\texttt{hash}, d \leftarrow \texttt{lhs}.\text{query}(\texttt{pred})$
            \STATE $\texttt{st}[\texttt{hash}] = [\texttt{terminal}]$
        \ENDFOR
        
        \STATE \textbf{return} $\texttt{st}$
    \end{algorithmic}
\end{algorithm}

\begin{algorithm}[tbh]
    \caption{hash\_simplify}\label{alg:hash_simplify}
    \begin{algorithmic}[1]
        \REQUIRE{ individual $\texttt{ind}$, simplification table $\texttt{st}$, lhs instance $\texttt{lhs}$, tolerance $\tau$ }
        \ENSURE{ simplified version of the individual $n$ }

        \FOR{$\texttt{subtree} \in \texttt{ind}$}
            \STATE $\texttt{pred} \leftarrow \texttt{subtree}.\text{predict}(\mathcal{X})$
            
            \IF{$\text{Var}(\texttt{pred}) = 0$}
                \STATE $\texttt{pred} \leftarrow \texttt{pred} \times 0.0$
            \ENDIF
            
            \STATE $\texttt{hash}, d \leftarrow \texttt{lhs}.\text{query}(\texttt{pred})$
    
            \IF{$\texttt{hash} \in \texttt{st} \text{ and } d \leq \tau$}
                \STATE $\texttt{st}[\texttt{hash}] \leftarrow \texttt{st}[\texttt{hash}] \cup \{\texttt{subtree}\}$
                \STATE $\texttt{subtree} \leftarrow \text{argmin} \text{ size}(\texttt{tree}) \text{ for } \texttt{tree} \in \texttt{st}[\texttt{hash}]$
            \ELSIF{$\texttt{hash} \not\in \texttt{st}$}
                \STATE $\texttt{lhs}.\text{index}(\texttt{pred})$
                \STATE $\texttt{st}[\texttt{hash}] = [\texttt{subtree}]$
            \ENDIF
        \ENDFOR
        
        \STATE \textbf{return} $\texttt{ind}$
    \end{algorithmic}
\end{algorithm}

When simplifying the expression, we can traverse the tree either top-down or bottom-up since we apply the replacements on the fly.
The order of traversal can generate different simplifications.
The advantage of the top-down approach is that it prunes large subtrees first (if it finds a hash entry in the first levels). This may lead to fewer nodes being visited as it may stop at an earlier level. 
On the other hand, the bottom-up traversal may require more steps, as a node simplified at the bottom level may trigger a new simplification at the upper level. 
Nevertheless, this fine-grained simplification may lead to larger simplifications after simplifying the smaller branches.

This simplification will not always return an algebraically equivalent expression but rather an approximation. By imposing a maximum distance threshold, we mitigate this issue by guaranteeing that this procedure will replace subtrees with similar semantics. This is an advantage for practical applications since we favor simpler expressions at the expense of minor differences in the predictions.

\section{Methods}~\label{sec:methods}

We implemented a standard evolutionary algorithm using the DEAP framework \cite{DEAP_JMLR2012}. The individuals are randomly initialized using the PTC2 method \cite{PTC2}. Then, iteratively, we perform a fixed number of generations of tournament selection with a tournament size of $3$. Possible variations operators are the crossover and mutations \textit{insert node}, \textit{remove node}, \textit{replace node}, and \textit{replace subtree}.

Three variants were implemented: \textit{without simplify}; \textit{bottom up}, and \textit{top down}. 
Simplification is done on the initial population and every offspring generated by the variation operators --- this way, we guarantee that every individual is simplified at least once. The individuals also go into the nonlinear parameter optimization method Levenberg-Marquardt \cite{levenberg1944method, marquardt1963algorithm} algorithm using Scipy \cite{2020SciPy-NMeth}, which was shown to be effective in previous SR algorithms \cite{ITWithCoeffsOptimization, ParameterIdentificationForSR, PrioritizedGrammarEnumeration}. After simplifying, we repeat the parameter optimization.

Our hash method is probabilistic (due to plane initialization); thus, the hash size is set to avoid collisions. We tried to increase the hash size by powers of two until the simplification table would not have any collision at its initialization step. During initial experiments, we found that if the initial table has a hash collision for the features or constant, the simplification may replace features with a constant, which is undesirable unless a feature is almost constant.

The threshold was set to a fixed value based on the smallest train MSE error in preliminary experiments over all datasets and set to $0.01$, so it is one order of magnitude below all the train errors. Population size and number of generations were set so most runs would finish under a $3$ wall clock hours. Table \ref{tab:sr-hyperparameters} describes the hyper-parameters used in the experiments.

\begin{table}[tbh]
    \centering
    \caption{Symbolic regression algorithms hyper-parameters.}
    \label{tab:sr-hyperparameters}
    \begin{tabular}{ll}
    \hline
    \textbf{Parameter}     & \textbf{Value}                                      \\ \hline
    pop size ($S$)     & $80$                                        \\
    max gen  ($G$)     & $200$                                       \\
    max depth ($\text{max}_d$)    & $7$                                         \\
    max size  ($\text{max}_s$)    & $128$ ($2^7)$                                        \\
    tolerance ($\tau$) & $1e-2$ \\
    hash\_len  & $256$ bits                                        \\
    probabilities & $1/5$ for each variation operator \\
    objectives    & [error (MSE), size (\# nodes))] \\ 
    Function set  & [$+$, $-$, $*$, $\frac{\cdot}{\cdot}$, $|\cdot|$, $\cos^{-1}$, $\sin^{-1}$, $\tan^{-1}$,\\& $\cos$, $\sin$, $\tan$, $e^{(\cdot)}$, $\mathtt{min}$, $\mathtt{max}$, \\& $\log$, $\log{(1+ \cdot)}$, $\exp{(1+\cdot)}$, $\sqrt{|\cdot|}$, $(\cdot)^2$] \\ \hline
    \end{tabular}
\end{table}

We used a selection of datasets to analyze the effect of simplification throughout the evolutionary process. Table \ref{tab:datasets} shows the name and dimensionality of the datasets used. We should note that our approach is agnostic to GP implementation and the choice of LSH. We have chosen the basic implementations to highlight the effects of such simplifications.

\begin{table}[tbh]
    \centering
    \caption{Dimensionality of the six datasets used to perform an in-depth analysis.}
    \label{tab:datasets}
    \begin{tabular}{lll}
    \hline
    \textbf{Dataset}        & \textbf{\# samples} & \textbf{\# features} \\ \hline
    Airfoil        & $1503$       & $5$           \\
    Concrete       & $1030$       & $8$           \\
    Energy Cooling & $768$        & $8$           \\
    Energy Heating & $768$        & $8$           \\
    Housing        & $506$        & $13$          \\
    Yacht          & $308$        & $6$           \\ \hline
    \end{tabular}
\end{table}

The loss function is the mean squared error (MSE) between the predictions $\mathbf{\widehat{y}} = f(\mathcal{X})$ and observed values $\mathbf{y}$:

\begin{equation}
    \text{MSE}(\mathbf{\widehat{y}}, \mathbf{y}) = \frac{1}{d} \sum_{i=1}^{d} \left ( \widehat{y}_i - y_i \right )^2.
\end{equation}

As a measurement of simplicity, we also use the concept of complexity by La Cava et al. \cite{la_cava_learning_2019}, defined for a node $n$ with $k$ arguments as the recursive combination of complexity of children with its root/head node: 

\begin{equation}
    C(n) = c_n * (\sum_{a=1}^k C(a)).
\end{equation}

By assigning high complexity values to hard-to-interpret operations, we can measure ---recursively--- how this propagates through the tree. The complexity for the operators is described in Table \ref{tab:complexity}.

\begin{table}[tbh]
    \centering
    \caption{Complexity of each operator.}
    \label{tab:complexity}
    \begin{tabular}{@{}ll@{}}
    \toprule
    \textbf{Complexity} & \textbf{Operators} \\ \midrule
    
    2          & $+$, $-$, $\mathtt{cte}$       \\
    
    3          & $*$, $\mathtt{max}$, $\mathtt{min}$, $(\cdot)^2$, $|\cdot|$      \\
    
    4          & $\frac{\cdot}{\cdot}$, $\sqrt{|\cdot|}$,  $e^{(\cdot)}$     \\ 
    5 &  $\exp{(1+\cdot)}$, $\cos$, $\sin$, $\tan$ \\
    6 &  $\cos^{-1}$, $\sin^{-1}$, $\tan^{-1}$ \\
    8 & $\log{(1+ \cdot)}$ \\
    9 & $\log$ \\\bottomrule   
    \end{tabular}
\end{table}

Each method was run 30 times with different split seeds for each dataset on the same hardware. The data was divided into three partitions: $50\%$ as train (visible to the algorithm to perform the parameter and function optimization); $25\%$ as validation (used to assess the loss during the evolution, but not used during train); and $25\%$ test (held-out data used to obtain the final values for the experiments). The validation split is also used to pick the final model returned by the algorithm.

Statistical comparisons, when reported, use the non-parametric Wilcoxon test with Holm-Bonferroni correction, and all comparisons made are explicitly depicted in the figures. Table \ref{tab:p-value-annotations} shows the annotations used to show statistical significance in the plots. When an annotation is shown as ``* (ns)'', it shows a statistical significance of one asterisk, but after correction, it becomes non-significant.

\begin{table}[tbh]
    \centering
    \caption{p-value annotations and their correspondences, after applying the alpha correction.}
    \label{tab:p-value-annotations}
    \begin{tabular}{ll}
        \hline
        Annotation & p-value                                    \\\hline
        ns  & $5\times 10^{-2} < \text{p} \leq 1.0$ \\
        *          & $1\times 10^{-2} < \text{p} \leq 5\times 10^{-2}$ \\
        **         & $1\times 10^{-3} < \text{p} \leq 1\times 10^{-2}$ \\
        ***        & $1\times 10^{-4} < \text{p} \leq 1\times 10^{-3}$ \\
        ****       & \hspace{5.2em} $\text{p} \leq 1\times 10^{-4}$ \\\hline
    \end{tabular}
\end{table}

All data and the source code for implementations, experiments, and post-processing analysis are available at \url{https://github.com/gAldeia/hashing-symbolic-expressions}.

\section{Results and Discussion}~\label{sec:results}

In this section we analyze different aspects of the experiment's results. First, we verify whether the simplification accelerates the convergence of GP to a good local optimum; next, we compare the number of simplifications performed by each traversal strategy; we show the distribution of the average size, complexity, and goodness-of-fit of the final solutions. We also show the relative difference of the solutions under these same criteria when paired with the individual runs. Finally, we compare the differences in runtime and explore some hand-picked simplification rules generated throughout the runs.

\subsection{Convergence}

\begin{figure}[tbh]
    \centering
    \includegraphics[width=\linewidth]{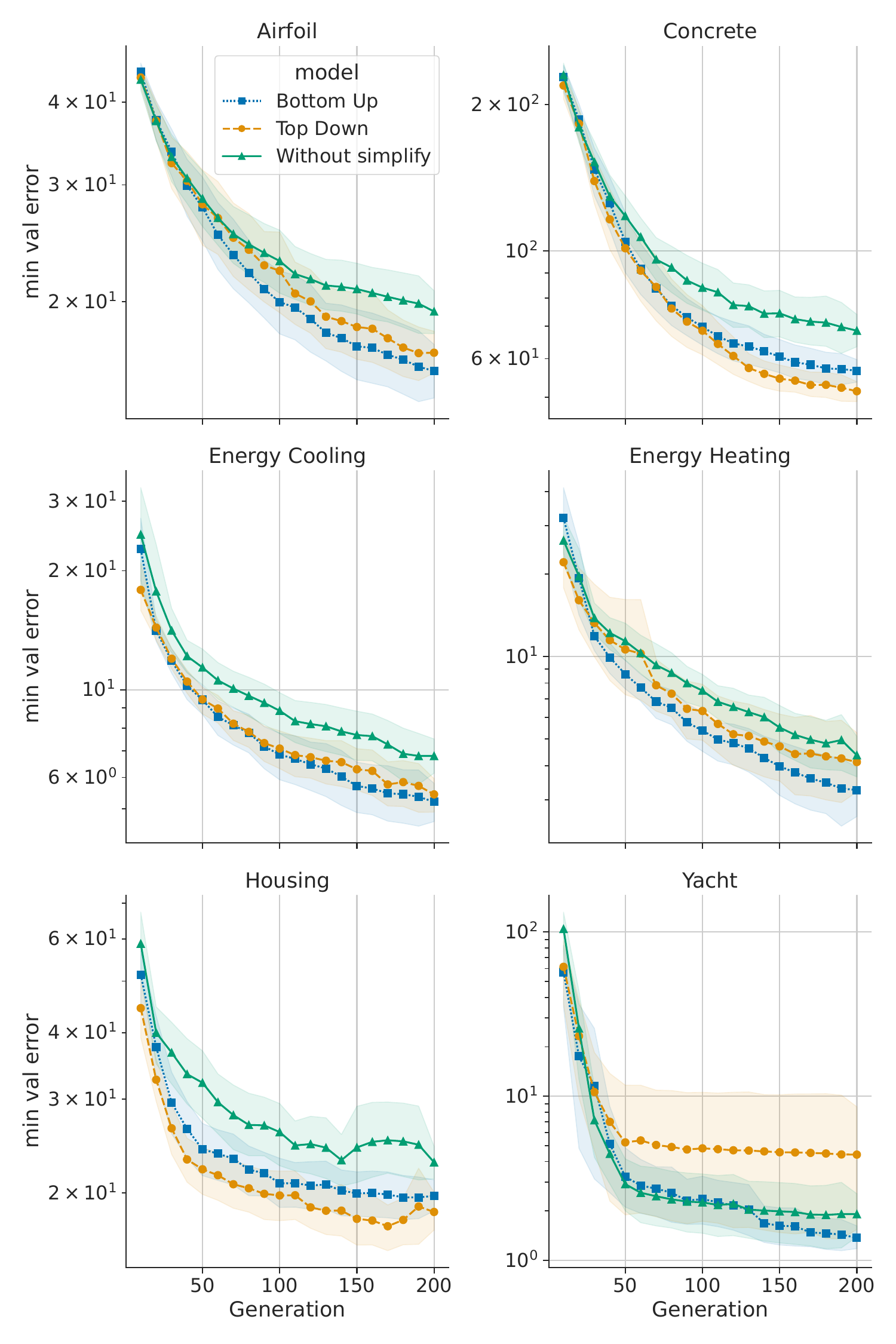}
    \caption{Validation error of the best individual during the evolution. $y$ axis is in log-scale.}
    \label{fig:convergence_error}
\end{figure}

The data was split into training and validation sets. While the model used the training partition to perform parameter and model optimization, the validation partition was used to compute the MSE with unseen data. Figure \ref{fig:convergence_error} reports the minimum validation error during the evolution, estimated using the mean of 30 runs.

These results show that the versions with simplification often outperform the traditional GP throughout the run, leading to a better local optimum. The only exception is the Yacht dataset, in which the top-down strategy performs worse than the bottom-up and the version without simplification. In every other dataset, both traversal strategies present similar performance.

\begin{figure}[tbh]
    \centering
    \includegraphics[width=\linewidth]{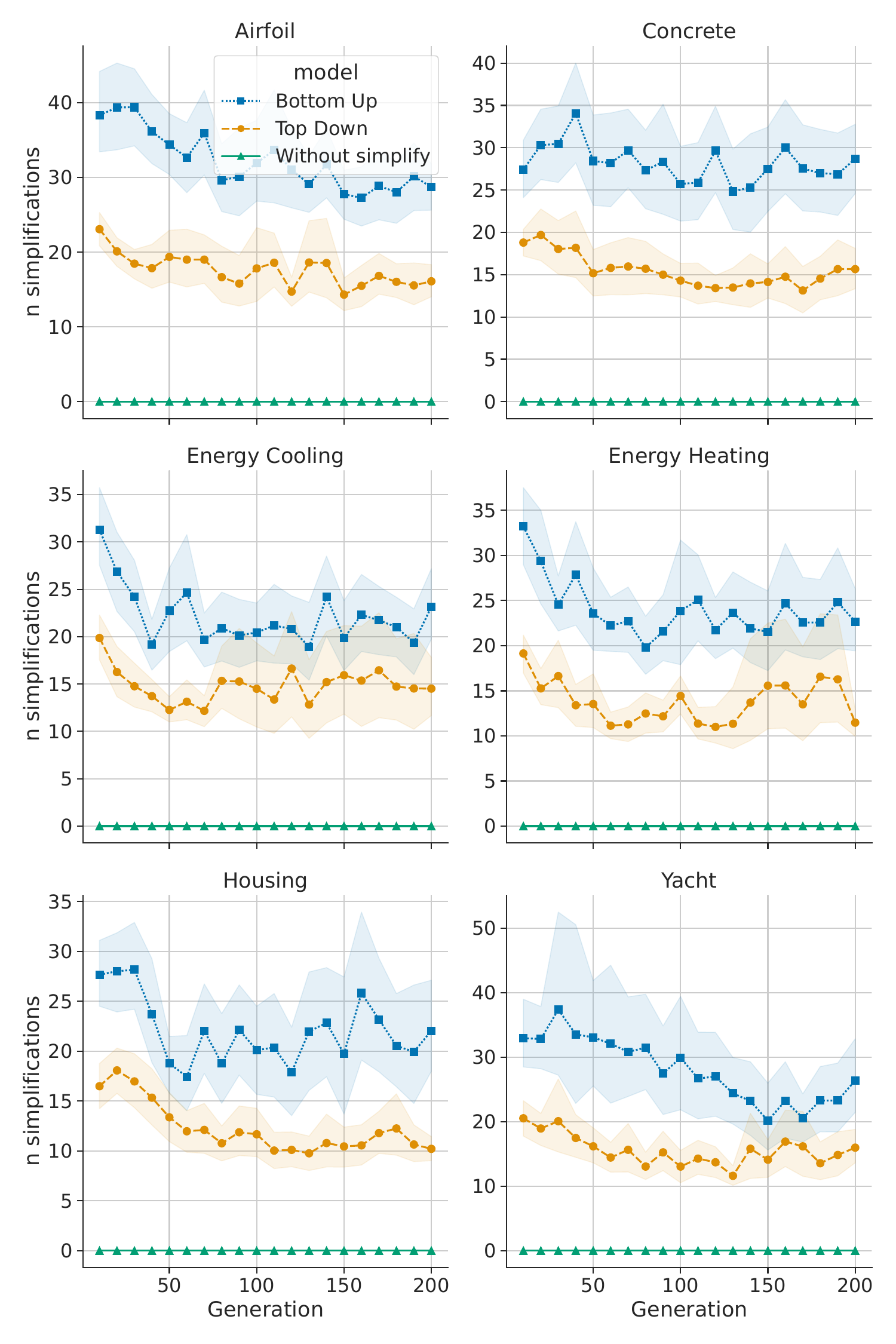}
    \caption{Number of simplifications performed in each generation.}
    \label{fig:n_simplifications}
\end{figure}

The two traversal strategies, as conjectured before, perform a different number of simplifications during the evolutionary process, as we can see in Figure \ref{fig:n_simplifications}.
The bottom-up strategy performs around $50\%$ more simplifications with a little intersection in the estimated confidence interval.
We believe that the number of simplifications appears to diminish in later generations as we effectively remove bloated and redundant subtrees; however, we also start to have more specialized and complex models, which are harder to simplify.

\subsection{Goodness-of-fit and size trade-off}

\begin{figure}[tbh]
    \centering
    \includegraphics[width=\linewidth]{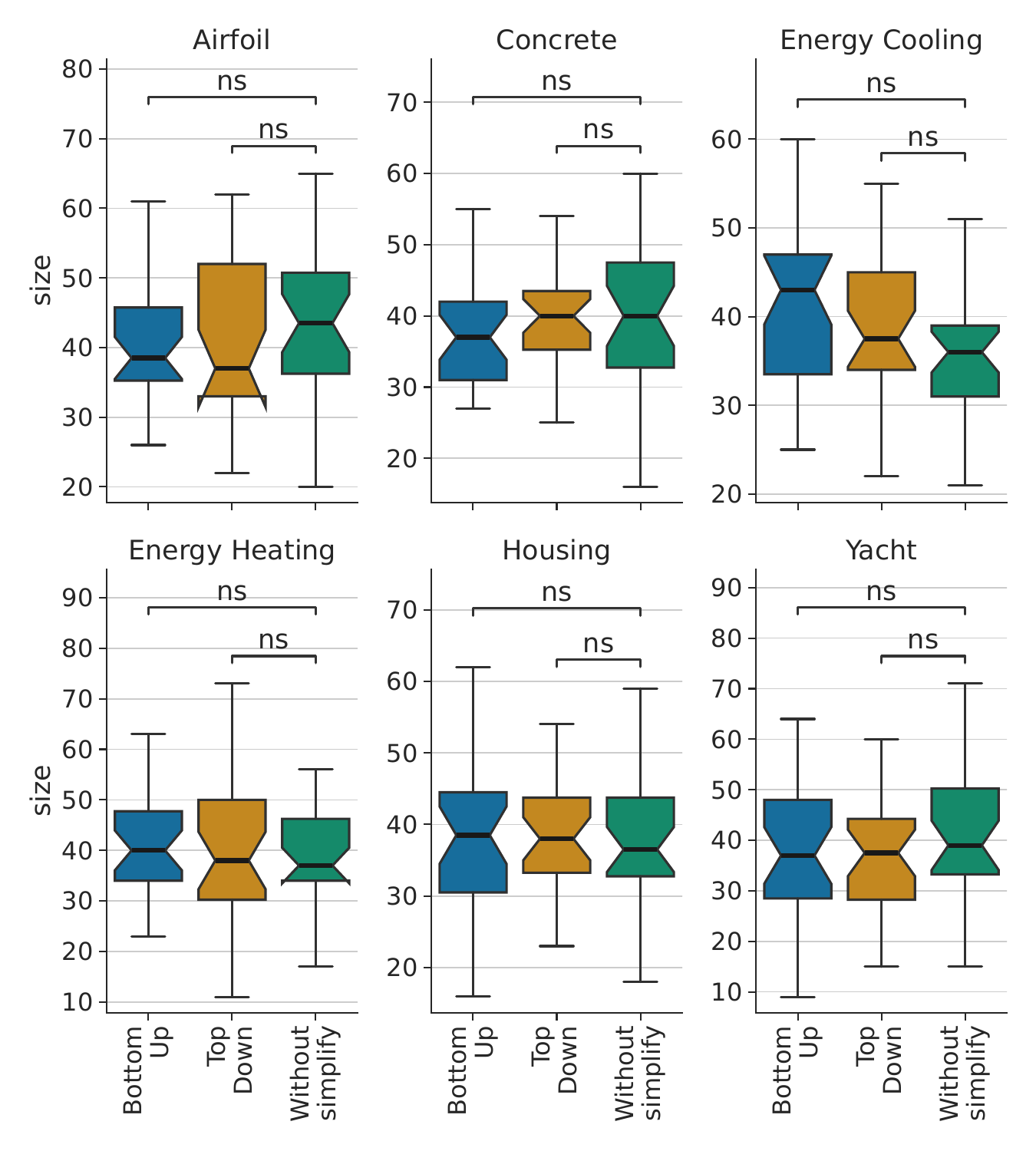}
    \caption{Final size of the solutions found by each method. The simplification methods showed no statistically different significance related to without any simplification.}
    \label{fig:size_boxplot}
\end{figure}

\begin{figure}[tbh]
    \centering
    \includegraphics[width=\linewidth]{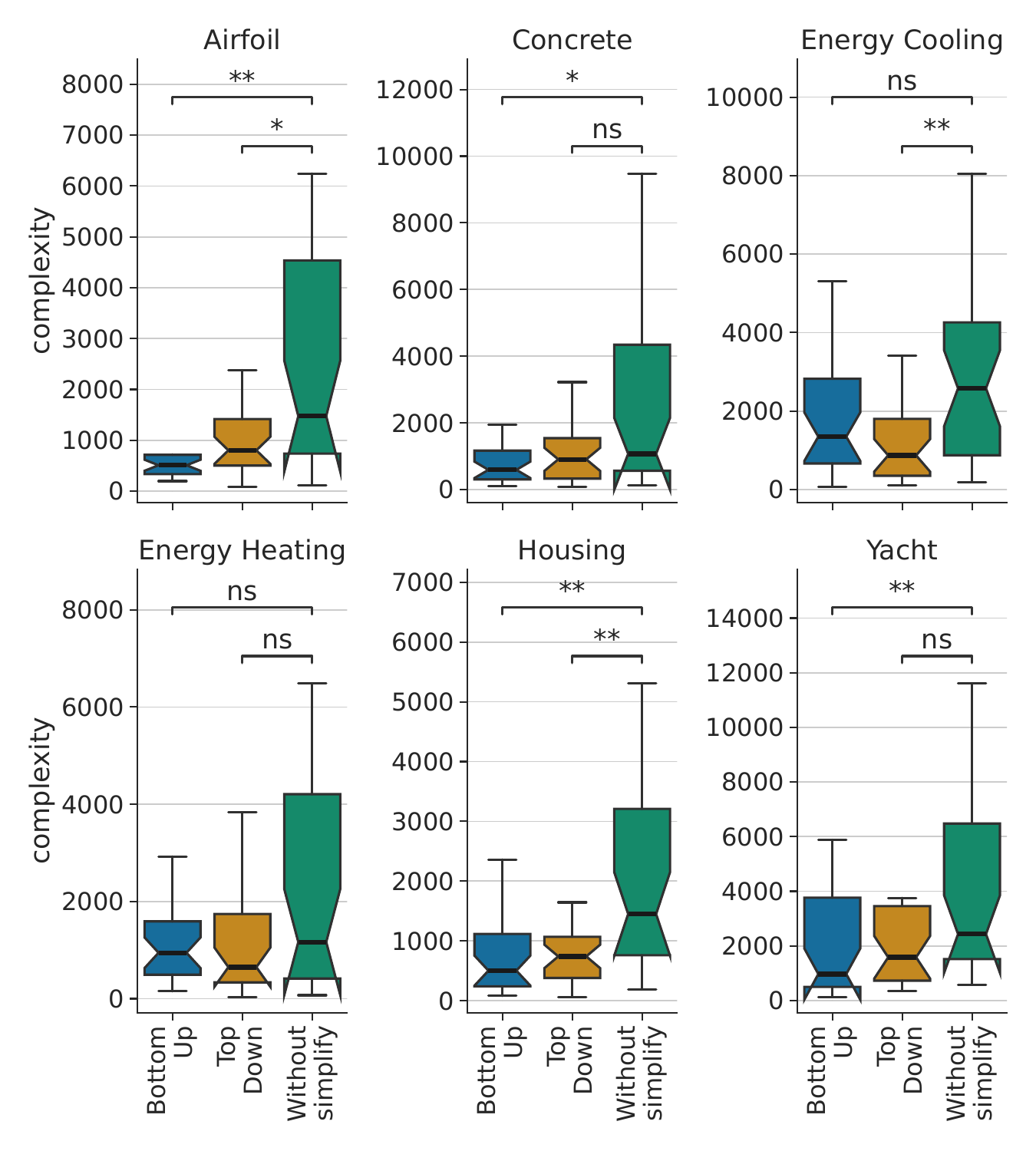}
    \caption{Final complexity of the solutions. There are several cases where simplification can improve complexity.}
    \label{fig:complexity_boxplot}
\end{figure}

\begin{figure}[tbh]
    \centering
    \includegraphics[width=\linewidth]{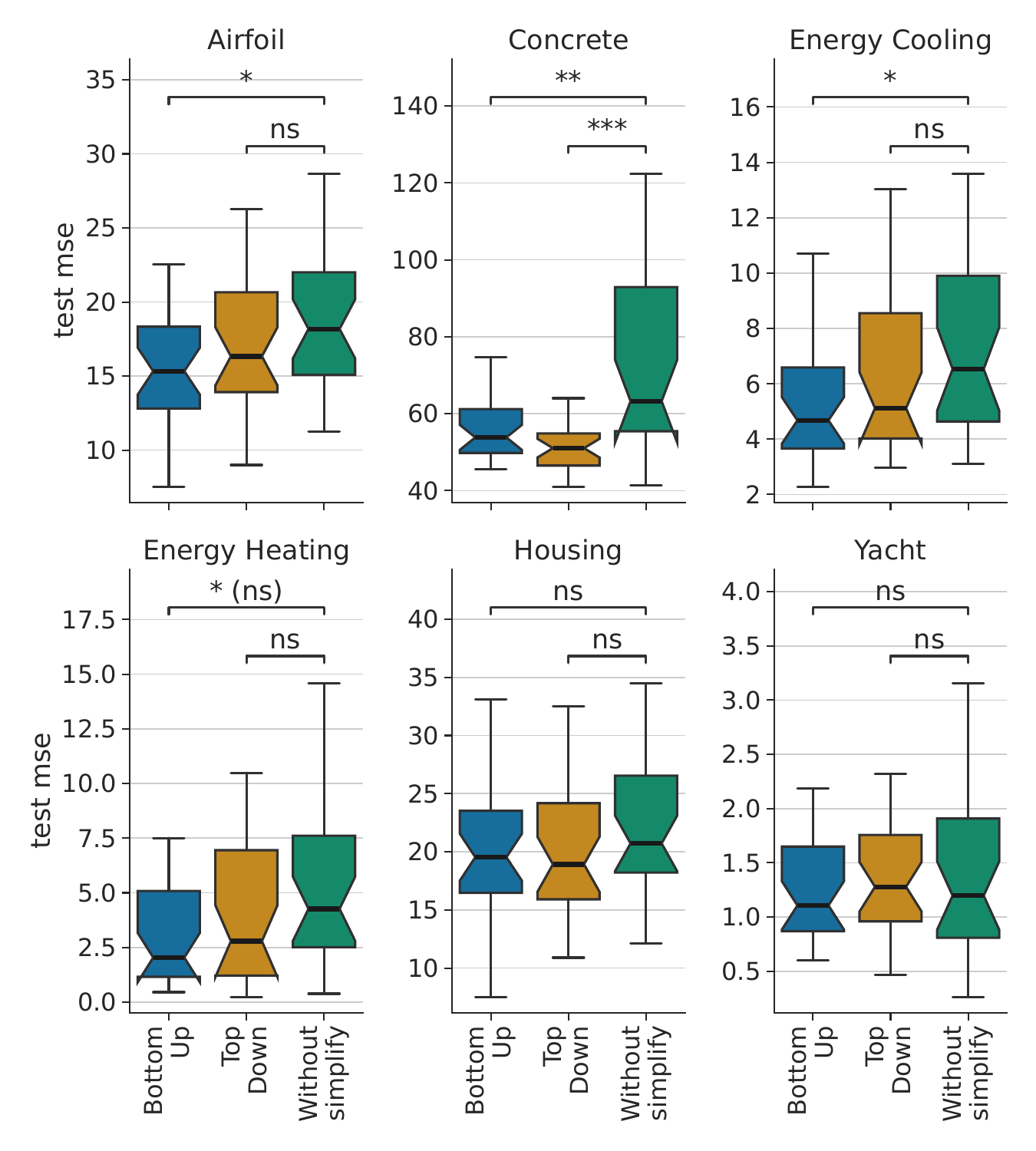}
    \caption{MSE on the test partition. Simplification led to better performance in error, with less complex models.}
    \label{fig:mse_boxplot}
\end{figure}

At the end of the run, every algorithm picks the final model with the best performance on the validation split. Figures \ref{fig:size_boxplot}, \ref{fig:complexity_boxplot}, and \ref{fig:mse_boxplot} report the size, complexity, and MSE on the test partition, respectively.

Regarding the size, we cannot reject the null hypothesis that there are not differences between the approaches --- a counter-intuitive result, as the simplification is expected to reduce the size of the expressions. However, looking at the complexity boxplots, we observe a smaller variation (for the best) using simplification strategies and the rejection of the null hypothesis with a p-value between $[10^{-3}, 10^{-2}]$. As the models are simplified during the evolution, it frees up space (within the limit of maximum nodes) to accommodate more useful sub-expressions. As a by-product, the generated expressions are also less complex when considering the definition of recursive complexity. We also hypothesize that the simplification opens up space for further improving the expressions, and the evolutionary process efficiently takes care of that by generating expressions of the same size but with better performance.

We find differences in complexity for the datasets Airfoil, Concrete, Energy Cooling, Housing, and Yacht. Even though they are equivalent in size, the simplification process led to finding solutions with better complexity without explicitly trying to minimize it.
Note that the simplified expression does not exhibit higher errors than the original. 
Regarding the test set MSE, Airfoil, Concrete, and Energy Cooling datasets show statistically significant improvements in the bottom-up strategy.
The top-down strategy shows improvements only for Concrete.

The simplification strategy shows only small differences and no clear better option in terms of the results, except for the number of simplifications. The bottom-up strategy will always iterate through the entire tree, while the top-down strategy can cut large branches right away, ending up with a smaller number of simplifications. 

We believe that Yacht was the most challenging dataset due to the test MSE error scale shown in Fig. \ref{fig:mse_boxplot}, as it is closer to the threshold than any other algorithm. Parameter optimization, especially the inexact simplification threshold, seems important to be adjusted based on the error scale, and we plan to investigate it further.

\subsection{Relative change}

\begin{figure}[tbh]
    \centering
    \subfloat[\label{delta_size}]{%
       \includegraphics[width=0.32\linewidth]{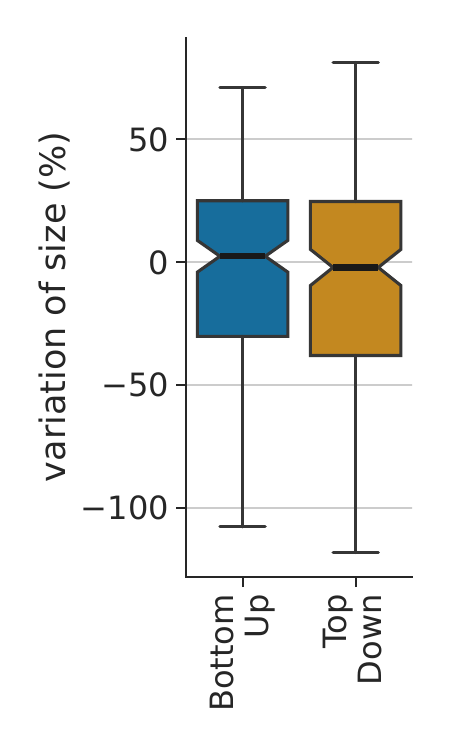}}
    \subfloat[\label{delta_complexity}]{%
       \includegraphics[width=0.32\linewidth]{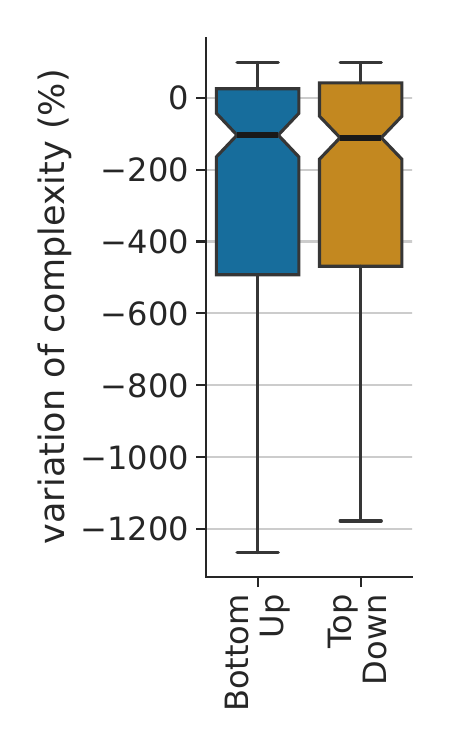}}
    \subfloat[\label{delta_test_mse}]{%
       \includegraphics[width=0.32\linewidth]{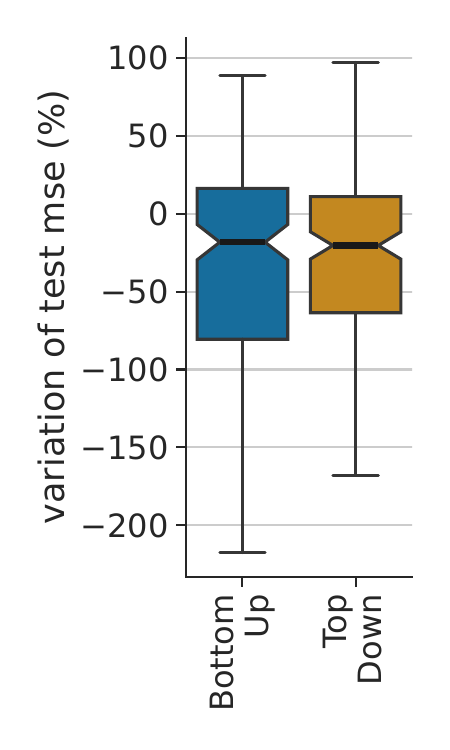}}
    \caption{Percentage variation of size (\ref{delta_size}), complexity (\ref{delta_complexity}) and MSE on test (\ref{delta_test_mse}) compared to without any simplification.}
    \label{fig:deltas}
\end{figure}

Even though the differences in MSE seem modest, those are considering the mean and median of the distribution. A better way to verify the benefits of the simplification strategy is to pair the $30$ runs based on their seed and calculate the percentage of variation between the simplification methods and without any simplification (i.e., the baseline). This way, the variation between datasets can be aggregated, and an overall measurement and a one-sided $t$-test can be applied to verify whether the mean of the relative change is different from zero. Figure \ref{fig:deltas} reports the variations for the bottom-up and top-down strategies.

A t-test for the mean of the distributions was performed for each subfigure with $180$ degrees of freedom ($6$ datasets and $30$ runs each).

Regarding size, we see the distribution with a median of $2.43\%$ for bottom-up and $-2.13\% $ for top-down, but both presented a p-value greater than $0.05$, so we cannot reject the null hypothesis.
Complexity shows a median of $-104.06\%$ and $-111.34\%$, with p-values of $2.94 \times 10^{-7}$ and $1.59 \times 10^{-6}$ for bottom-up and top-down, respectively. This indicates that for each individual run, we observe a reduction in complexity compared to the baseline.
For MSE, the medians are $-18.21\%$ and $-20.32\%$, and the p-values are $8.165 \times 10^{-7}$ and $1.85 \times 10^{-5}$. Again, we can reject the null hypothesis and conclude the simplification reduces the error by $20\%$ on average.

\subsection{Analysis of hashes and individuals}

We performed a single run on the Yacht dataset to obtain a few insights on how simplification is replacing nodes. We chose it because it has no statistically significant differences in size and error and is also low-dimensional (for the sake of the example). 
At the end of the run, we had a total of $5144$ hash entries created from $7324$ expressions. 
This subsection will review some of the entries of the simplification table. An entry is a tuple of a hash and a list of equivalent trees. The hash will be truncated to simplify the discussion, and expressions will be shown as string-formatted versions of the expressions, ordered by smallest to largest --- meaning that any of the subtrees would be replaced by the first.

Let us take a look into the first case:

\begin{Verbatim}[frame=single]
1010110111001101...
 - square(x_5)
 - multiply(x_5, x_5)
 - absolute(square(x_5))
 - maximum(square(x_5), x_0)
 - maximum(add(-15.455, x_1), square(x_5))
\end{Verbatim}

We first notice that, without explicit rules, our method started to replace \verb|multiply(x_5, x_5)| by \verb|square(x_5)|, a shorter representation.
It also learned that the absolute value of the square is always positive, so \verb|absolute(square(x_5))| could also be simplified to \verb|square(x_5)|.
Some simplifications there are based on data: that $x_5^2$ always dominates $x_0$ in the $\mathtt{max}$ function. Algebraic rules would not capture this simplification.

\begin{figure}[tbh]
    \centering
    \includegraphics[width=1\linewidth]{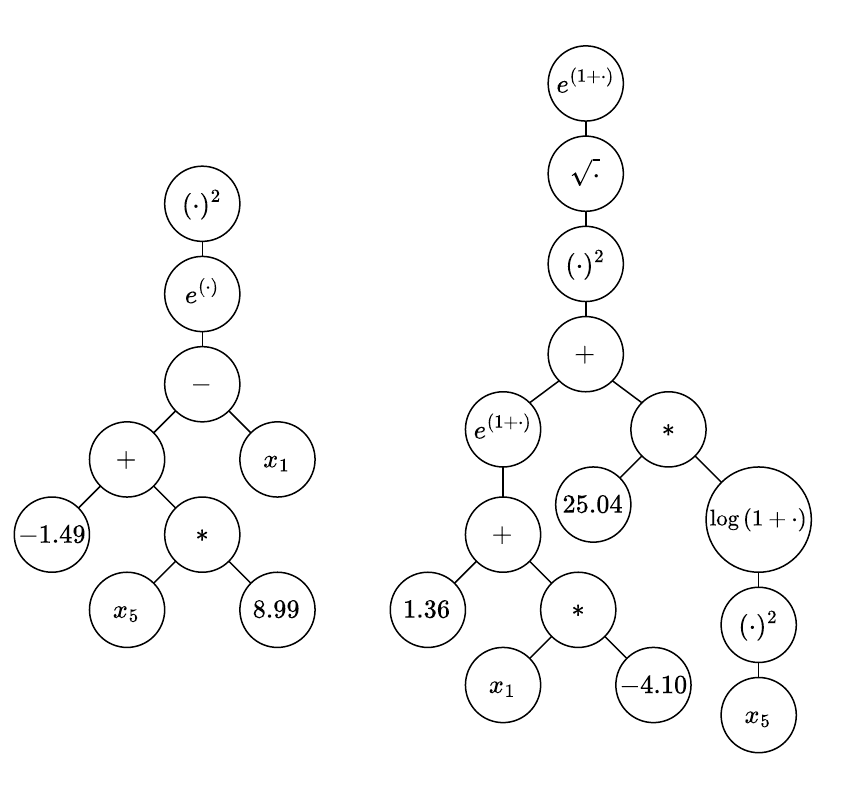}
    \caption{Smallest expression found by bottom-up (left) and without simplification (right) for the Yacht dataset. The MSE for each expression was $1.98$ using the bottom-up simplification and $2.08$ without simplification.}
    \label{fig:example_expressions}
\end{figure}

Some other interesting cases arise. Here, we have a permutation of arguments on a $4$-ary commutative operation, as well as the chain of commutative operations:

\begin{Verbatim}[frame=single]
1110001011011111...
 - multiply(x_1, x_7, x_0, x_4)
 - multiply(x_0, x_4, x_1, x_7)
 - multiply(x_1, x_0, multiply(x_4, x_7))
\end{Verbatim}

Finally, some other cases are the simplification of redundant operations (as in \verb|minimum(x_2, x_2)|), the usage of the identity value of an operation as argument (as in \verb|add(0.0, x_2)|), and the chaining of $f$ with its $f^{-1}$ (as in \verb|log(exp(x_2))|):

\begin{Verbatim}[frame=single]
1010101111001001...
 - x_2
 - absolute(x_2)
 - minimum(x_2, x_2)
 - minimum(x_2, x_6)
 - minimum(x_2, x_5)
 - add(0.0, x_2)
 - log(exp(x_2))
 - sqrtabs(square(x_2))
 - maximum(-523.249, x_2)
\end{Verbatim}

Figure \ref{fig:example_expressions} shows the smallest expression found by the bottom-up strategy and with GP without simplification for the Yacht dataset.

The bottom-up model obtained an MSE on the test set of $1.98$ with $9$ nodes, and GP without simplification obtained a test error of $2.08$ using $15$ nodes.
We can see more complex constructs when not applying simplification, as well as the chaining of $\sqrt{\cdot}$ and $(\cdot)^{2}$.

The yacht dataset has errors that are one order of magnitude greater than the simplification threshold, and besides not showing improvements in error, it had improvements in complexity.
By analyzing one single run, we proved that the benefits of our simplification methods extend beyond size and error. The method was able to minimize complexity without being explicitly told to do so.

\subsection{Execution time}

\begin{figure}[tbh]
    \centering
    \includegraphics[width=\linewidth]{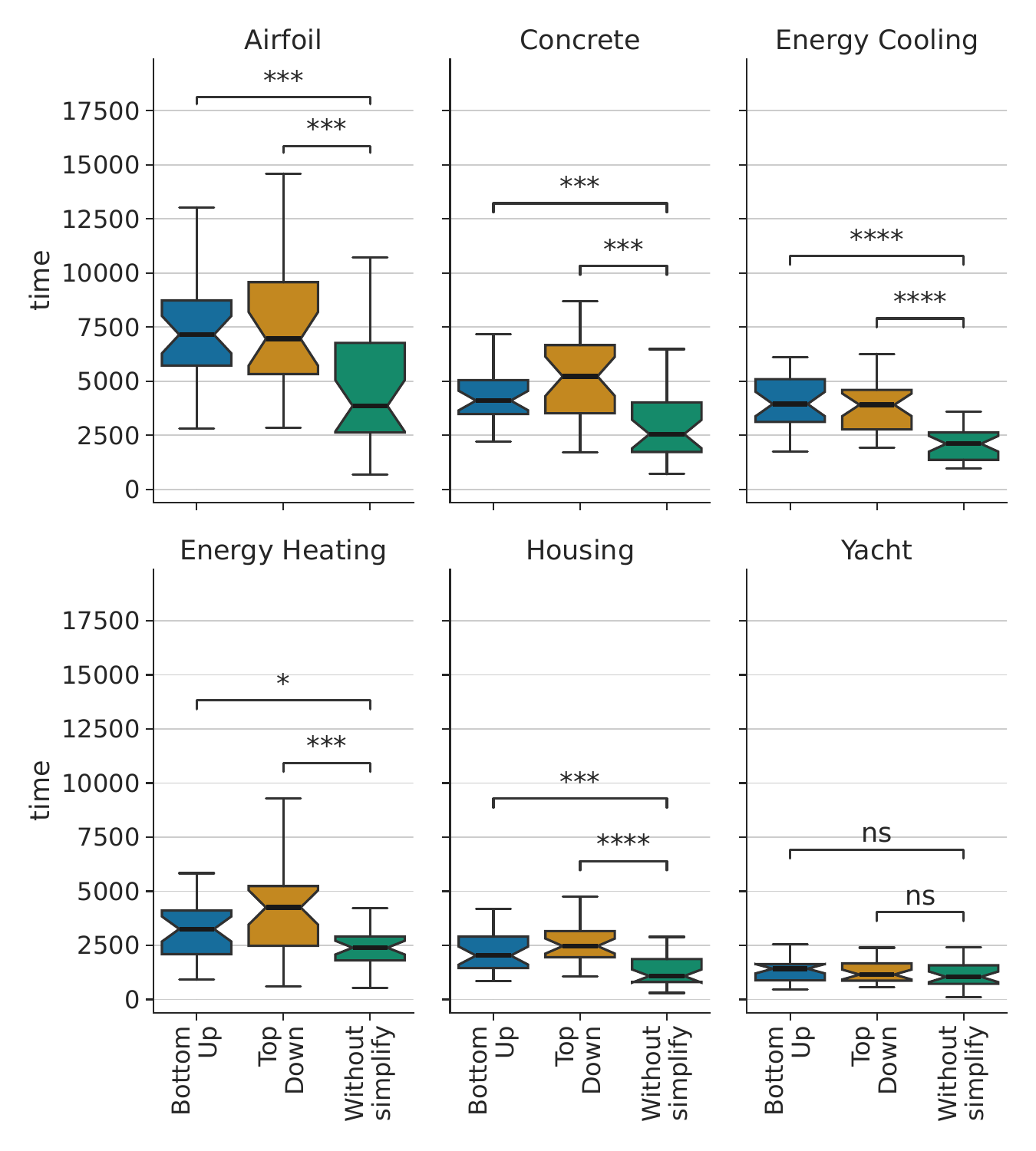}
    \caption{Variation of execution time (in seconds) for each method. The subplots are sharing the $y$ axis.}
    \label{fig:time_boxplots}
\end{figure}

Figure \ref{fig:time_boxplots} reports the execution time for each algorithm.

We notice that the simplification methods take more time to run than without any simplification, with statistical differences for all but the last dataset.
This happens because we perform a nonlinear optimization twice for every individual, as they are re-optimized after the simplification, indicating that the method would benefit from caching recent information to avoid re-optimizing the entire tree after every simplification.

\section{Conclusions}~\label{sec:conclusions}

In this paper, we propose an inexact simplification method using LSH to learn substitution rules that are either well-known algebraic identities or simplifications specific to the dataset domain. LSH efficiently stores and queries expressions that behave similarly in the phenotypical space, enabling us to apply the simplification process for every sampled expression throughout the evolution.

With this approach, we can experimentally analyze the influence of simplification and bloat control on the evolutionary process. The empirical results showed that, on average, the versions with simplification returned more accurate expressions of the same size but with less complex construct. Comparing runs departing from the same seed, we observed a median of $20\%$ reduction in the mean squared error when using simplification.

This approach has some drawbacks. For example, the hash function can only guarantee that similar expressions are clustered with high probability. We can eventually include expressions with distinct behavior in the wrong cluster, but this should be rare. Another issue is that the simplification is inexact, so it can perform substitutions that slightly change the behavior of the function or that are exact only on the training data.

On the other hand, since this is a data-driven approach, it can learn the rules on the fly without any need to pre-determine the algebraic identities. It can also learn rules specific to the dataset (e.g., $x_1$ is always smaller than $x_2$). 

As for the next steps, we intend to improve the influence of simplification by making a ``warm-up'' step, where random individuals are generated and used to initialize the table; we will also evaluate the influence of performing simplification sporadically, only for a selection of individuals, or only at the final generation; and how does the threshold affect the results. We will also test some other hash functions with better guarantees than SimHash. Finally, we will integrate this approach into a high-performance GP implementation to make it possible to run more detailed tests, also including other simplification methods into the benchmark.

\begin{acks}
W.G.L. was supported by National Institutes of Health (NIH) grant R00-LM012926, and Patient Centered Outcomes Research Institute (PCORI) ME-2020C1D-19393.
F.O.F. is supported by Funda\c{c}\~{a}o de Amparo \`{a} Pesquisa do Estado de S\~{a}o Paulo (FAPESP) grant 2021/12706-1, and Conselho Nacional de Desenvolvimento Cient\'{i}fico e Tecnol\'{o}gico (CNPq) grant 301596/2022-0.
G.S.I.A. is supported by Coordena\c{c}\~{a}o de Aperfei\c{c}oamento de Pessoal de N\'{i}vel Superior (CAPES) finance Code 001 and grant 88887.802848/2023-00.
\end{acks}

\bibliographystyle{ACM-Reference-Format}
\bibliography{sample-base}

\end{document}